\documentclass[letterpaper, 10pt, conference]{ieeeconf}      

\IEEEoverridecommandlockouts                              

\usepackage{graphics} 
\usepackage{epsfig} 
\usepackage{mathptmx} 
\usepackage{times} 
\usepackage{amsmath} 
\usepackage{amssymb}  
\usepackage{url} 
\let\labelindent\relax
\usepackage{enumitem}
\setlist{leftmargin=3.5mm}
\usepackage{footnote}
\usepackage{flushend}

\title{\LARGE \bf
LoCoQuad: A Low-Cost Arachnoid Quadruped Robot for Research and Education
}

\author{Manuel Bernal$^{1}$ and Javier Civera$^{2}$
\thanks{*This work was not supported by any organization}
\thanks{$^{1}$Manuel Bernal Lecina is with School of Engineering and Architecture, University of Zaragoza, Spain
        {\tt\small mbernallecina@gmail.com}}%
\thanks{$^{2}$Javier Civera Sancho is with the RoPeRT - Robotics, Perception and Real Time Group, University of Zaragoza, Spain
        {\tt\small jcivera@unizar.es}}%
}

\begin{document}

\maketitle
\thispagestyle{empty}
\pagestyle{empty}

\begin{abstract}

Developing real robotic systems requires a tight integration of mechanics, electronics and software. Most of the times, existing robotic platforms are either closed or expensive or both, and in-house solutions are costly to develop and maintain. Open-source and low-cost designs are essential to facilitate the access to real robotic platforms and enable further progress in the field.

LoCoQuad (see Fig. \ref{fig:teaser}) is an arachnoid quadruped platform that we designed targeting research and education in robotics. To meet these two ends, our platform allows for a high degree of flexibility and configurability. Our legged design has the lowest hardware cost of the state of the art, in the range of 150-165USD. We validated the robot platform by running several experiments showing over all functionalities. All the mechanical and electronic designs and all the software have been made open source and can be found at \url{https://github.com/TomBlackroad/LoCoQuad}

\end{abstract}

\section{INTRODUCTION}

The robotic ecosystem has experienced a significant transformation in the last decade. Robotics was in its beginnings an academic discipline, with a promising future but a wide array of scientific challenges. Today, there are already several commercial applications of robotic technologies, and even more of them are supposed to be a reality in a short- and mid-term future and are object of heavy investment. Among others we can name vacuum cleaners, autonomous cars or service robots.

Such an exciting progress in robotic science and systems has occurred due to a combination of factors, such as increasing computational power and advances in perception, learning, control or mechanics. However, in spite of such fast developments, it is essential to keep a critical view and be aware of the scientific challenges that still lie ahead. Among them, we believe three are very relevant in relation to the availability of robotic platforms and systems.

\begin{itemize}

\item Robotics research is many times focused on very specific sub-problems and \textbf{lacks integration and evaluation in fully autonomous robotic platforms}. For example, research on perception is most of the times done without robots \cite{mur2015orb}, and research on control uses very frequently off-board perception, e.g. MoCap systems \cite{mellinger2012trajectory}. Integrating perception and action might cross-fertilize both topics and open new research areas with huge potential, as discussed for example in \cite{bohg2017interactive}. While several platforms already exist in the market, their cost and lack of flexibility limit their use in robotic research.

\item \textbf{Open-source, low-cost, flexible and easy-to-use platforms would also allow fair comparisons} between different algorithms. Research on robot learning, for example, would be more widely affordable for researchers and practitioners. Notice how one of the reasons of the boost of machine learning and computer vision research has been the existence of public widely recognized datasets (\emph{e.g.}, \cite{lecun-mnisthandwrittendigit-2010,deng2009imagenet}) that allow rigorous comparisons.

\item \textbf{Education on robotics depends strongly on the availability of platforms} that are easy to use, but containing tools, software and hardware that are as close as possible to those of real robots. Flexibility and low cost are also relevant aspects. Most of the existing platforms lack such flexibility, have a closed design, or use ad-hoc hardware that limit the generalization of the learned competences.

\end{itemize}

\begin{figure}[!t]
      \centering
      \includegraphics[width=\linewidth]{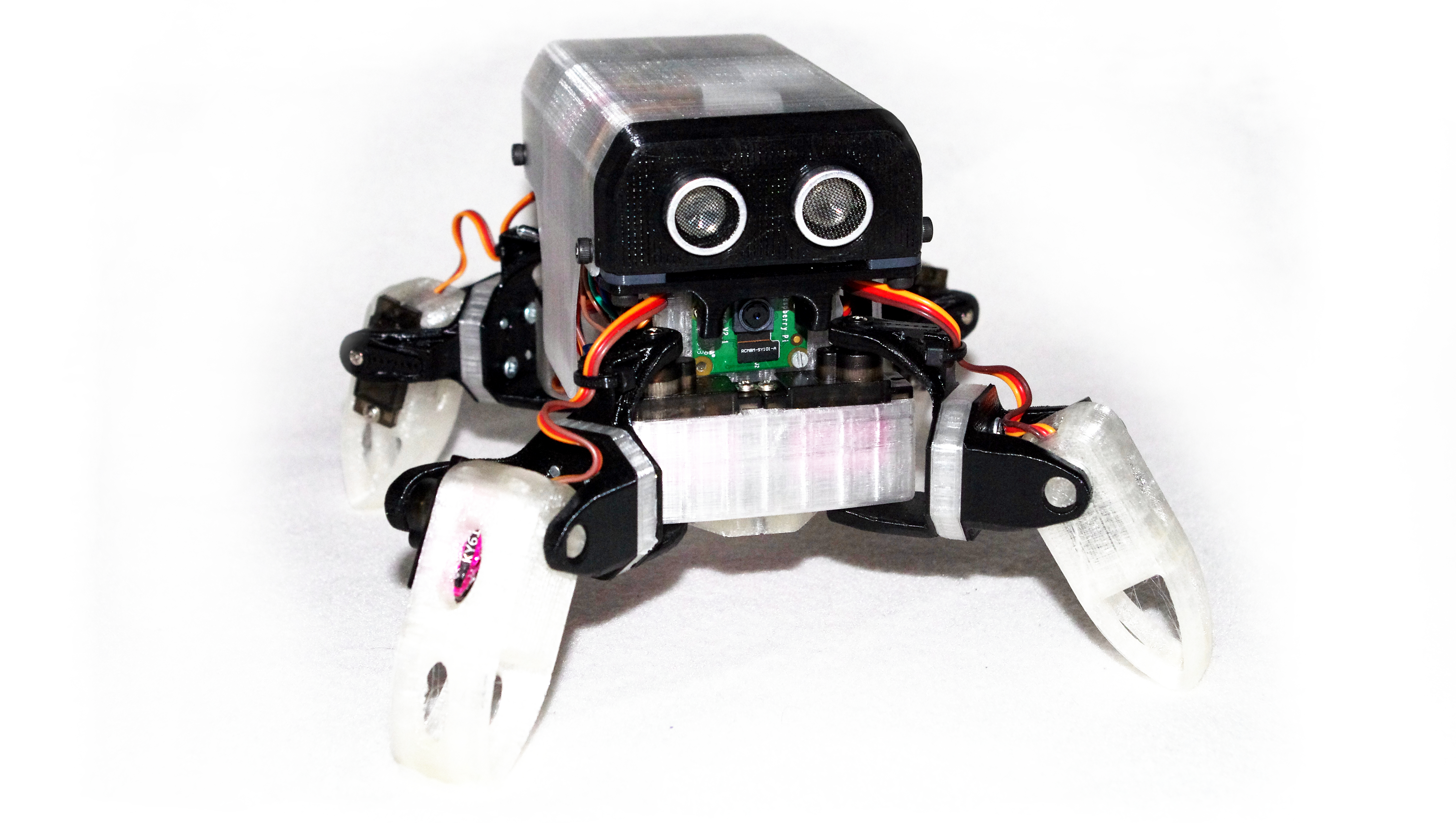}
      \caption{The LoCoQuad robotic platform.}
      \label{fig:teaser}
\end{figure}

In this paper we present LoCoQuad, an open-source low-cost and general-purpose robotic platform targeting robotic research and education. Its development was motivated by the need of low-cost flexible platforms in wich we could implement state-of-the-art algorithms in reinforcement learning. There are several options in the market that allow for some degree of flexibility (e.g. the Lynxmotion kits\footnote{\url{http://www.lynxmotion.com/}}), but none of them are low-cost. Our design is based on standard components and can be easily 3D-printed, so that it can be self-made with a total cost in the range of 150USD (to our knowledge, the lowest in the state-of-the-art for a functional computer-based legged robot). We chose a Raspberry Pi as its computational unit, in order to be flexible and maintainable in the future. This implementation offers important advantages over microcontrolled robots, due to its flexibility and usability for researchers and students lacking an electronic background. We provide examples of several basic tasks in the paper and the web, validating the design and highlighting its potential.


\section{RELATED WORK}

\begin{savenotes}
\begin{table}[!h]
\caption{LoCoQuad COST vs Other ROBOTIC PLATFORMS}
\label{tab:cost_comparison}
\begin{center}
\begin{tabular}{|l|c|c|}
\hline
\textbf{Platform} & \textbf{Price [\$]} & \textbf{Robot type}\\
\hline
\hline
Aracna~\cite{lohmann2012aracna} & 1389 & Quadruped\\
\hline
E-Puck~\cite{mondada2009puck} & 280 & Wheeled\\
\hline
Phantom X MRIII\footnote{\url{https://www.trossenrobotics.com/phantomx-ax-hexapod.aspx}} & 1300 & Hexapod\\
\hline
Roomba~\cite{tribelhorn2007evaluating} & 200 & Wheeled \\
\hline
CotsBots~\cite{bergbreiter2003cotsbots} & 200 & Wheeled  \\
\hline
Kilobot~\cite{rubenstein2012kilobot} & 100 & Vibration\\
\hline
Khepera IV~\cite{mondada1999development} & 2670 & Wheeled\\
\hline
PiArm & 263 & Arm\\
\hline
NAO power V6~\cite{gouaillier2009mechatronic} & 9000 & Humanoid\\
\hline
Cozmo\footnote{\url{https://anki.com/en-us/cozmo.html}} & 180 & Wheeled\\
\hline
Sphero RVR\footnote{\url{https://www.sphero.com/rvr}} & 250 & Wheeled\\
\hline
RoboMaster S1\footnote{\url{https://www.dji.com/es/robomaster-s1}} & 400 & Wheeled\\
\hline
Molecubes~\cite{zykov2007molecubes} & 350 & Modular\\
\hline
Lynxmotion SQ3U & 550 & Quadruped\\
\hline
\textbf{LoCoQuad} & \textbf{150-165} & Quadruped\\
\hline
\end{tabular}
\end{center}
\end{table}
\end{savenotes}

\subsection{Closed and/or Expensive Platforms}

The number of robotic platforms that are low-cost and open-source has been traditionally very low \cite{piperidis2007low}. The vast majority of robotic platforms available for research and education are closed, meaning that parts of their mechanical or electronic designs are not available and cannot be re-configured \cite{fujita1998development,raibert2008bigdog}. Their high cost also limits a widespread use in the research and education communities, although their technical performance is many times impressive.

Another trend in the community is using slightly modified or available commercial platforms, for example the Roomba in \cite{tribelhorn2007evaluating}. Although the price in this case is lower than in other platforms (see Table~\ref{tab:cost_comparison}), the users are in most cases limited by the default robot configuration, and the details of the mechanical and electronic design might be closed.

\subsection{Low-Cost Platforms for Research and Education}

There is a wide array of platforms exclusively designed for education. \cite{lopez2016andruino} presents a wheeled platform based on Arduino, which limits its computation possibilities. \cite{zykov2007molecubes} is another interesting project, but lacking the generality of our design. Aracna \cite{lohmann2012aracna} or Khepera \cite{mondada1999development} are other interesting projects, but their unit prices are an order of magnitude higher than ours.

Research institutions sometimes develop their own in-house platforms \cite{yim2000polybot,5069841}. However, most of the times, the cost of such designs and the risk of obsolescence are high.

There are several projects aiming, like us, at low-cost solutions. Kilobot \cite{rubenstein2012kilobot} is a micro-robot that vibrates in order to move and has been used for swarm robotics. \cite{sibley2002robomote,mcmickell2003micabot} are also low-cost platforms for research on sensor networks. \cite{kamimura2001self} presents a robot focusing on reconfigurability. Contrary to us, all those platforms target a specific application domain and lack generality. \cite{tan2016bio} details a bio-inspired design for an arachnoid robot. Although very similar to ours, its flexibility and computation are more limited. 

In our case, we developed LoCoQuad aiming to build an accesible robot, not focussing on accuracy but configurability. To this point, we were extremely cautious to provide the maximum versatility at the lowest price, renouncing to implement joint feedbacks.

Table \ref{tab:cost_comparison} shows a comparison of the costs of several robotic platforms of the literature, some of them already addressed in this section. Notice that legged robots are a minority, and our design is the one with the lowest cost in such category. \cite{mclurkin2014robot} also contains a cost comparison between different platforms, confirming our views on the low number of legged platforms available and their high cost.

\section{LOCOQUAD OVERVIEW}

LoCoQuad is a general-purpose robotic platform capable of walking, turning, waving and swinging by actuating over its four legs. In the design presented in this paper we included several sensors: a frontal camera, an inertial measurement unit (IMU) with accelerometer and gyroscope and two ultrasonic depth sensors. As we use a \emph{Raspberry Pi} as control unit, the sensor set is flexible and can be extended.

In this section we present the details on the mechanical, electronic and software design (sections \ref{sec:mechanics}, \ref{sec:electronics} and \ref{sec:software} respectively).

\subsection{Mechanics}
\label{sec:mechanics}

\begin{figure}[!t]
      \centering
      \includegraphics[width=\linewidth]{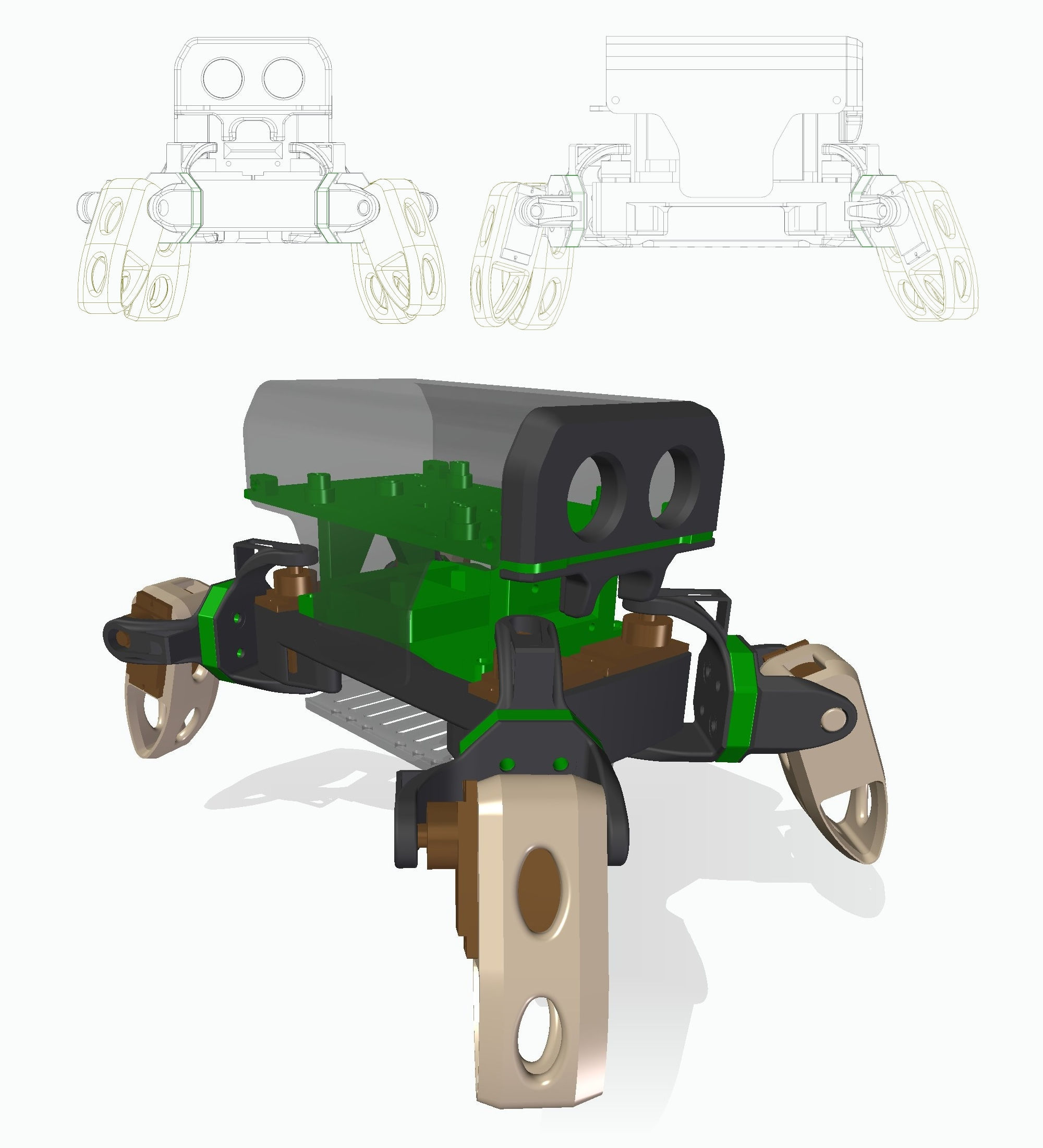}
      \caption{LoCoQuad basic configuration, with 2 joints per leg.}
      \label{fig:2joints}
\end{figure}

The mechanical structure of LoCoQuad is the one of a spider quadruped~\cite{kar2003design}.
All the electronics and sensors are placed in the main body. The body follows a stack design, being one of the design goals that the center of gravity is as low as possible, as it strongly affects the balance of the robot. To meet this goal, the battery pack is placed in the bottom stack, the power and driver boards in the middle stack, and the control board along with the sensors are in the top stack.

From the body, four legs are connected using commercial servo horns placed on custom 3D staples designed for that purpose. In order to connect each link of the legs, we have designed different 3D parts to fit different ranges of motion as well as a wide range of relative positions and configurations. In its most basic design with 2 joints per leg (rotator and knee), LoCoQuad has 4 different configurations per knee. In an alternative 3-joints-per-leg design (rotator, elevator and knee) the number of configurations is 16 due to the 4 at the elevator by the 4 at the knees (see Fig.~\ref{leg_configs}). These mechanical variations allow the platform to change and adopt different shapes, which is a relevant feature for research purposes.  

\begin{figure}[thpb]
      \centering
      \includegraphics[width=\linewidth]{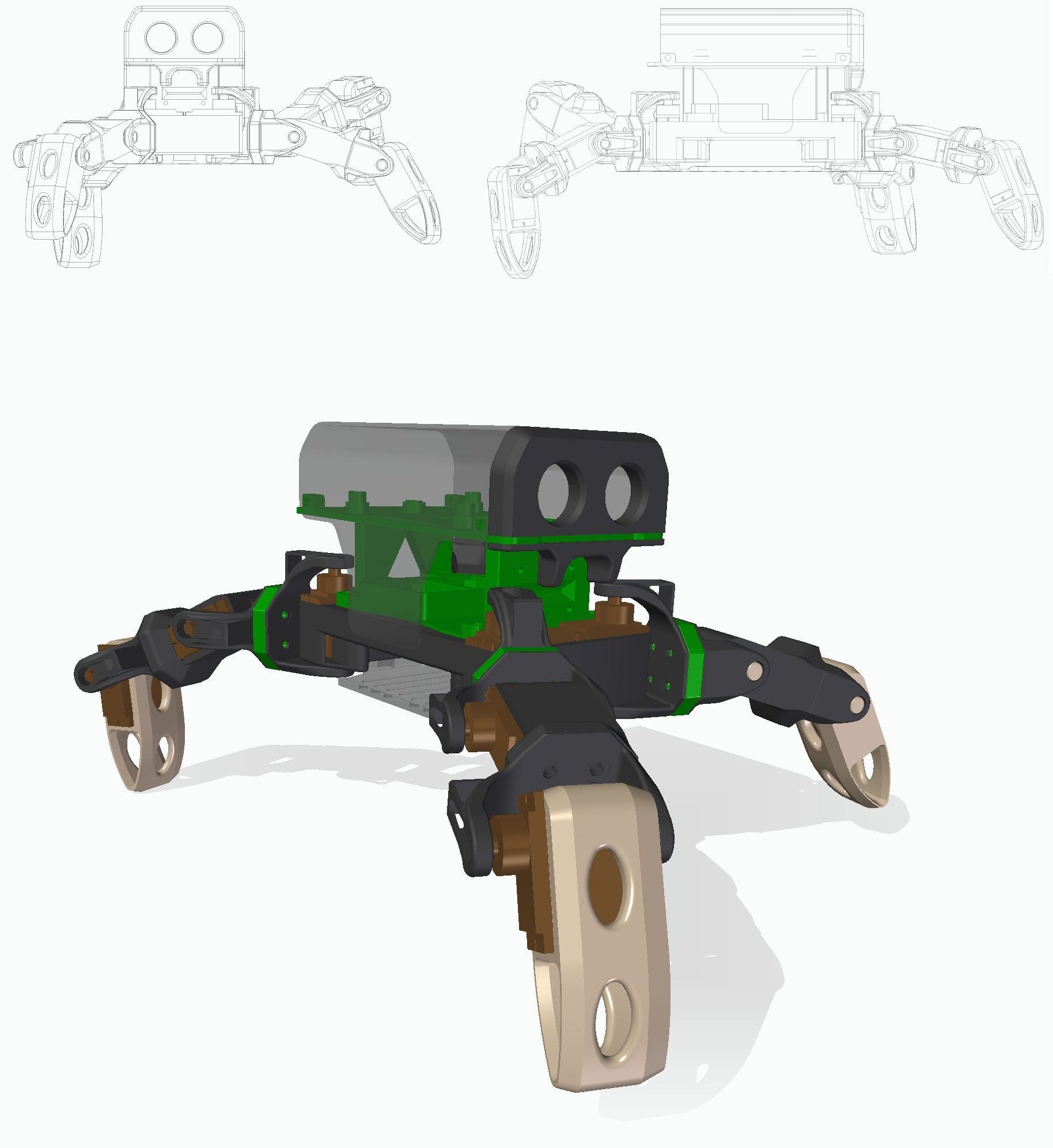}
      \caption{LoCoQuad alternative configuration, with 3 joints per leg.}
      \label{fig:3joints}
\end{figure}

Fig.~\ref{fig:2joints} shows the basic configuration for our LoCoQuad design, having 2 joints per leg. This is the configuration that we used for our first validation experiments. The high degree of flexibility of our platform, however, allows partial changes. In Fig.~\ref{fig:3joints} we show the alternative design that has 3 joints per leg.

\begin{figure}[thpb]
      \centering
      \includegraphics[width=\linewidth]{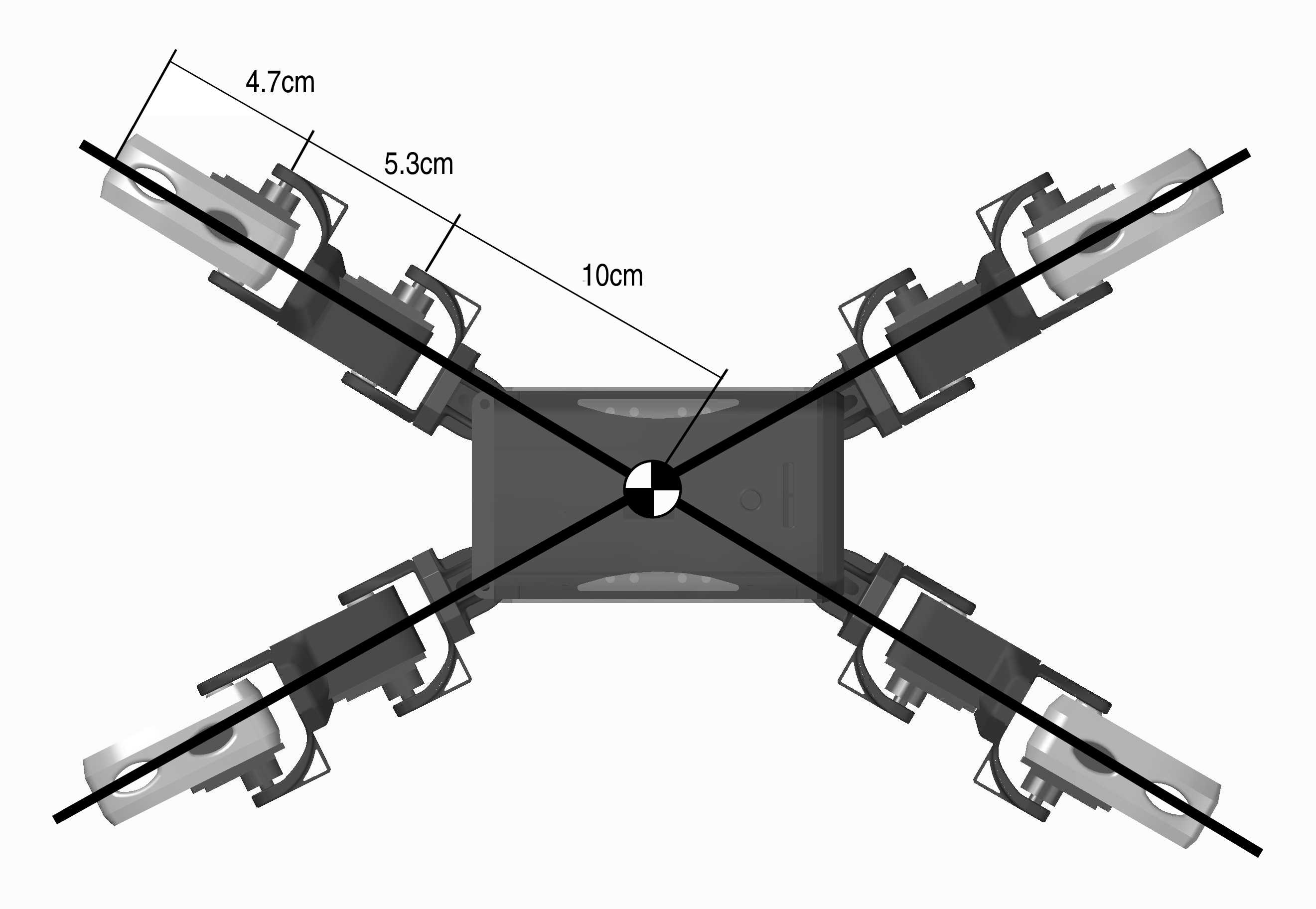}
      \caption{Worst case scenario for the torque analysis.}
      \label{fig:worstcasetorque}
   \end{figure}

One of the most critical aspects of the design was the size and shape of the robot, for the actuators to be able to handle the torque in the joints due to the robot weight. Since the aim was keeping the lowest possible cost, we chose the \emph{MG90S} servos as actuators, available for less than 5USD. Based on their nominal torque ($18$ N·cm), the length between joints was chosen to be $4.7$ cm for the end-effector, $5.3$ cm for the distance between elevator joint and knee joint and $10$ cm between center of gravity and elevator joint (see Fig. \ref{fig:worstcasetorque}). Since the robot has many potential configurations, the torque analysis was made for the worst case scenario (the one illustrated in Fig. \ref{fig:worstcasetorque}, with legs extended). 

Due to having 3D printed parts, the weight of links 2 and 3 (elevator and knee) change depending on the settings and the infill selected to print the parts. We consider a reference weight of 30 grams for these links, which is heavier than the real parts we use, as a safe value with a considerable margin.

\begin{figure}[thpb]
      \centering
      \includegraphics[width=\linewidth]{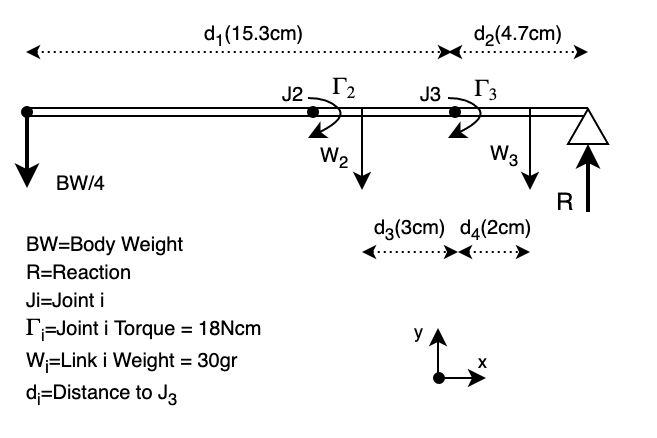}
      \caption{Force and torque diagram.}
      \label{fig:torque_diagram}
   \end{figure}

Applying equilibrium of forces ($\sum F \mid _{y} = 0$) and torques ($\sum \Gamma \mid _{J3} = 0$) we obtain the maximum weight for the robot body plus the load that can be carried.

The value of the reaction at the end-effector gives us

\begin{equation}
\label{ec:reaction}
R=\frac{BW}{4} + W_2+W_3
\end{equation}

Then, considering that both servos at elevator and knee are working at nominal torque, we have:

\begin{equation}
\label{ec:equilibrium}
\frac{\Gamma_2 + \Gamma_3}{g}=\frac{BW}{4}d_1+ W_2d_3-W_3d_4+Rd_2
\end{equation}

\noindent where $g$ is the gravity acceleration. 



The maximum weight $BW$ of the robot body, according to Eq.~\ref{ec:equilibrium}, is $850$ grams. The basic configuration with 2 joints per leg (Fig. \ref{fig:2joints}) has a total weight of $560$ grams, and the configuration with 3 joints per leg (Fig. \ref{fig:3joints}) a total weight of $670$ grams. It should be remarked that these limits are for the worst case configuration, that will not be reached in many applications. Both configurations, then, have a total weight well under the maximum, and the weight of the robot can be increased if needed. For example, they could carry small loads or be equipped with additional sensors or extra parts.

Concerning aesthetics, we also made an effort in our mechanical design for the robot to have a pleasant aspect. We believe that this is not critical for its use in research, but can help to introduce them in the lower educational levels. 

\subsection{Electronics}
\label{sec:electronics}

\begin{figure}[!ht]
      \centering
      \includegraphics[width=.9\linewidth]{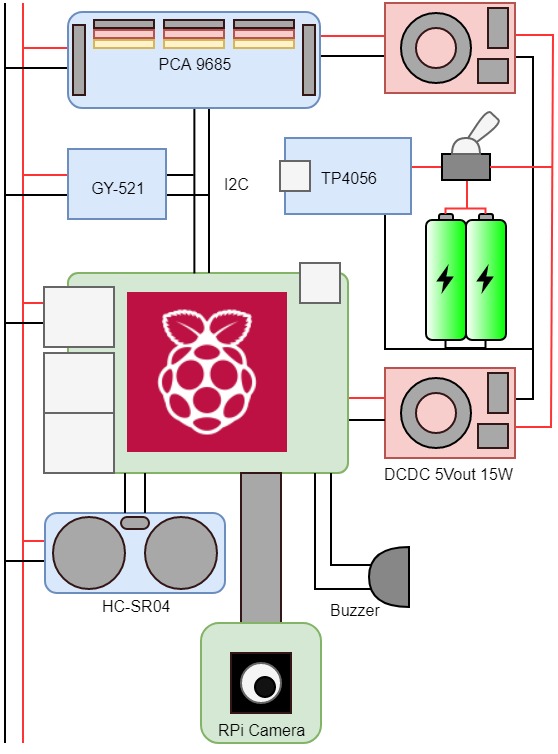}
      \caption{General electronics schematics.}
      \label{fig:sch}
\end{figure}

The electronic components of LoCoQuad are a control unit, driver boards to control the actuators and the sensors, power boards to provide enough current at desired voltages and a battery pack with enough capacity for the robot to move autonomously for a reasonable amount of time. Fig.~\ref{fig:sch} shows the electronic schematics, and the rest of the section describes each of the components.

Each one of these components has been selected and tested for a good compromise between performance, cost, power demand, availability and configurability. Starting with the open source control unit, a \emph{Raspberry Pi 3 Model B} has been chosen due to its connectivity, low power consumption and the versatility to execute different programming languages.

In order to control the high number of servos (\emph{MG90S}) that the robot might have (in our two designs we have 8 and 12 of them), we need a board with a high number of output channels. Based on this, the driver selected is a board carrying a \emph{PCA-9685} chip that allows to control up to $16$ Channels over the \emph{I2C} communication protocol. This protocol is managed by the control unit throughout its \emph{GPIO} pins. The board has a power closed circuit for the servos and a control circuit to enable the chips capabilities.

For each sensor a different interface is used. The IMU (\emph{GY-521}) is linked to the same \emph{I2C} bus as the servos driver. The buzzer and the depth sensor (\emph{HC-SR04}) are connected to the \emph{GPIO} pins of the \emph{Raspberry Pi}, since their control can be implemented directly into the control unit. And the Raspberry Camera module is connected to the integrated \emph{CSI} port. With this configuration, many pins and \emph{USB} ports are still available and could be used for alternative designs, tasks or experiments.

The \emph{Raspberry Pi} and the servo drivers have a $5V$ input and a high current drain. Hence, the output of the power boards should be at the same voltage or a little higher, and able to provide the desired current at any time.

Two \emph{DCDC} converters are used to provide this requirement. We chose LiPo batteries because of the high current drain they can provide at a high energy density. Simple 186500 LiPo cells give an output voltage between $3.7V$ and $4.2V$, which should be increased to $5.1V$. The \emph{Raspberry Pi} current peak is around 3A at its maximum amplitude in the most extreme cases. And the current drain from each servo is about $200mA$, giving a maximum current drain, for $12$ servos, of $2.4A$. Based on this, the power \emph{DCDC} converters need to output at least 27.5W (5.1V·(3A+2.4A)=27.5W). We choose to separate the output voltage of the actuators from the \emph{Raspberry Pi} one. This is because of the need to use low-cost and small \emph{DCDC} converters. In this case, we need two \emph{DCDC} converters from a $3.2V-4.2V$ input to a $5.1V$ output and a maximum current of $3A$. This allows each \emph{DCDC} converter to provide $15W$ (a total of $30W$), leaving us a safety margin in the power to drain more current if needed.

For the battery pack, a two parallel $3.7V$ LiPo cells were selected. Considering the high power consumption of the DCDC converters ($30W$), the batteries must be capable of a high and constant current drain. Depending on the capacity of the cells, the robot will have different usability life cycles. We selected an individual capacity of $3300$mAh. Assuming that the demand is the maximum ($30W$) and is kept constant over time (worst case scenario), from a battery at $3.8V$ average voltage the current drain would be almost 8A. Hence, our two parallel batteries, with a total capacity of $6600$mAh, would discharge in approximately \emph{36 minutes} (the losses of the \emph{DCDC} converters have been taken into account). It should be remarked that this analysis is for the worst case scenario: The consumption of the \emph{Raspberry Pi} is lower after booting and servos are not usually in operation all the time. 36 minutes is hence a lower bound for the autonomy of our design and our experiments have shown that for most use cases the batteries will last longer.

Targeting usability, in particular for educational use, we designed a charging module (\emph{TP4065}) capable of charging the LiPo battery at $1A$. In this manner our platform can be re-charged without removing the batteries. Also, to isolate the charging module from the rest of the system a toggle switch was installed. This allows two modes, \emph{Off/Charging} mode and \emph{On} mode.   

\subsection{Code}
\label{sec:software}

Although the core contributions of this paper lie in the mechanical and electronic design, we developed code for several basic functionalities, in order to test and validate the hardware (see Section \ref{sec:experiments} for details on the validation). 

We believe that the most important features to incorporate in these earliest stages of the platform design were drivers and basic structures. As we use open-source hardware, part of the drivers are based on open-source code, but others are our own developments. We chose to develop the basic functionalities for validation in \emph{Python}, as the \emph{Raspberry Pi GPIO} is nicely integrated with it and also provides easy interfaces for rapid prototyping and for our educational purposes. Migrating the code or programming directly in C++ would also be straightforward in the platform.

As we mentioned in \ref{sec:electronics}, the implementation of the Raspberry Camera module was an strategic step forward, giving extra capabilities to the system and allowing it to actually see and process the environment the system is runing into. Within the basic structure of the robot's code we provide a LoCoQuad class that includes finite state machines for different purposes such as initialization, rest, explore, picture shot or interactions. We combine them with specific data files containing parameter definitions, that can be easily edited to define all the possible LoCoQuad configurations. We think that this structure would be ideal to explain basic concepts and fundamentals on robotics, such us behaviour implementation, locomotion and perception. We also implemented motion primitives, that can also be modified or duplicated in order to achieve new objectives.

The basic robot motions and operations are controlled from the main class we developed, and can be included specifically into the \emph{explore state} the robot is after initializing actuators, sensors and communications. 

\section{EXPERIMENTS}
\label{sec:experiments}

\begin{figure*}[thpb]
      \centering
      \includegraphics[width=.85\linewidth]{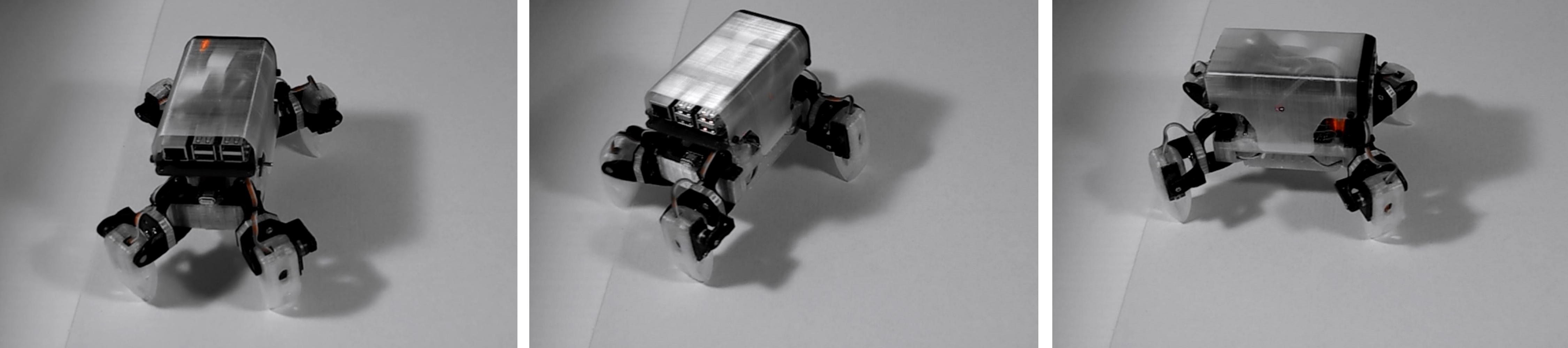}
      \caption{Turning motion validation.}
      \label{fig:turn_val}
\end{figure*}

\begin{figure*}[thpb]
      \centering
      \includegraphics[width=.85\linewidth]{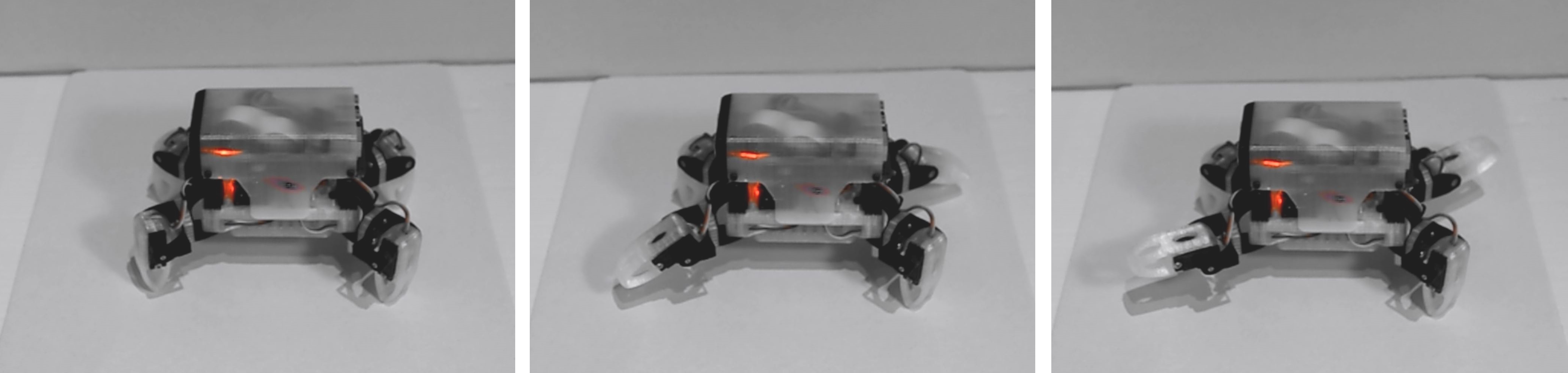}
      \caption{Balance control validation.}
      \label{fig:balance}
\end{figure*}

We have validated LoCoQuad at several stages of the project. First, during the design process, we built several units and made several iterations until we achieved the stable configuration presented on this paper. 

After we achieved a stable design, we developed code for testing the basic motion capabilities. Specifically, we developed functions for walking forwards and backwards and turning. Fig. \ref{fig:turn_val} shows three frames captured during a turning motion.

Finally, we performed several experiments combining sensing and action capabilities, in order to test the functionality of the robot for complex tasks. 

\begin{itemize}

\item \textbf{Sensor-based start-up routine}. We know if a user grabs the robot reading the IMU gravity direction. We use such information to start programmed routines, which is comfortable for the users.

\item \textbf{Photographer routine}. We implemented code for the robot to take pictures. The robot is first configured to look up to take better pictures.

\item \textbf{Obstacle avoidance}. We implemented a routine where the robot moves straight and, if an obstacle is detected by the ultrasonic sensor, the platform moves in a different direction in order to avoid it.

\item \textbf{Keeping balance}. We implemented a controller that, based on the gravity direction (read from the IMU), keeps the balance only on two legs (raising the other two). Fig. \ref{fig:balance} shows several pictures of the robot performing such experiment.

\end{itemize}

All these experiments can be better seen in the video accompanying the paper. We also created a YouTube channel\footnote{\url{https://www.youtube.com/channel/UCDpJfCahsBopnxOfMt28w4w}} where a high-resolution versions can be found. We will upload to the YouTube channel all our future developments.

\section{OTHER DESIGN DETAILS}

\begin{figure}[thpb]
      \centering
      \includegraphics[width=\linewidth]{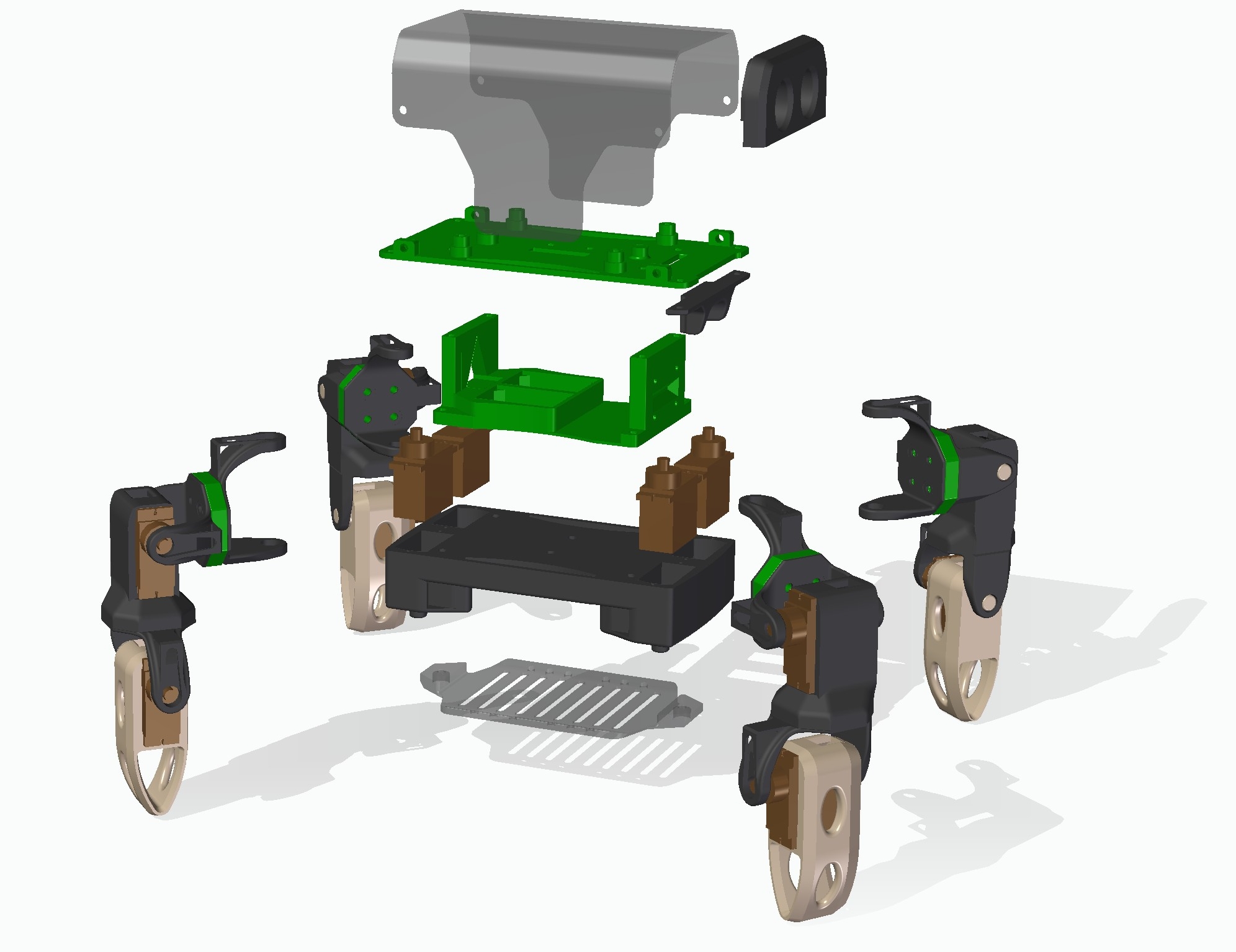}
      \caption{Assembly view.}
      \label{fig:assam}
   \end{figure}

The total size of the robot was designed based on two criteria: First, the robot components should fit into standard 3D printers, for its printing and assembly to be widely affordable. Secondly, we kept a small size for the robot in order to be easy to transport and store several units for multi-robot experiments. We also kept in mind the need for a simple assembly process.  In Fig.~\ref{fig:assam} we show the assembly view for the 3D printed parts in perspective.

\begin{figure}[!ht]
      \centering
      \includegraphics[width=\linewidth]{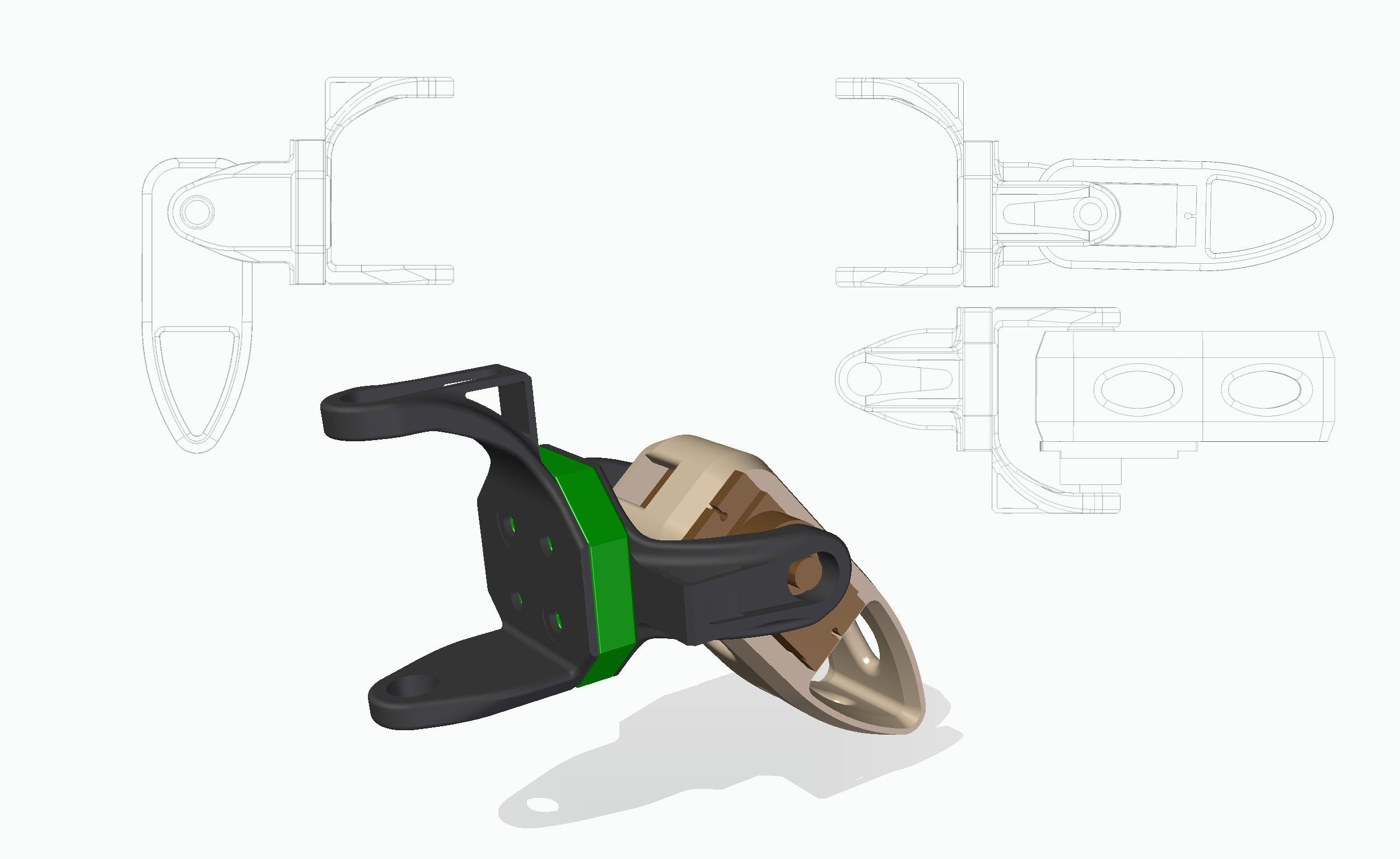}
      \caption{Two joints leg design.}
      \label{fig:2jointleg}
   \end{figure}

\begin{figure}[!ht]
      \centering
      \includegraphics[width=\linewidth]{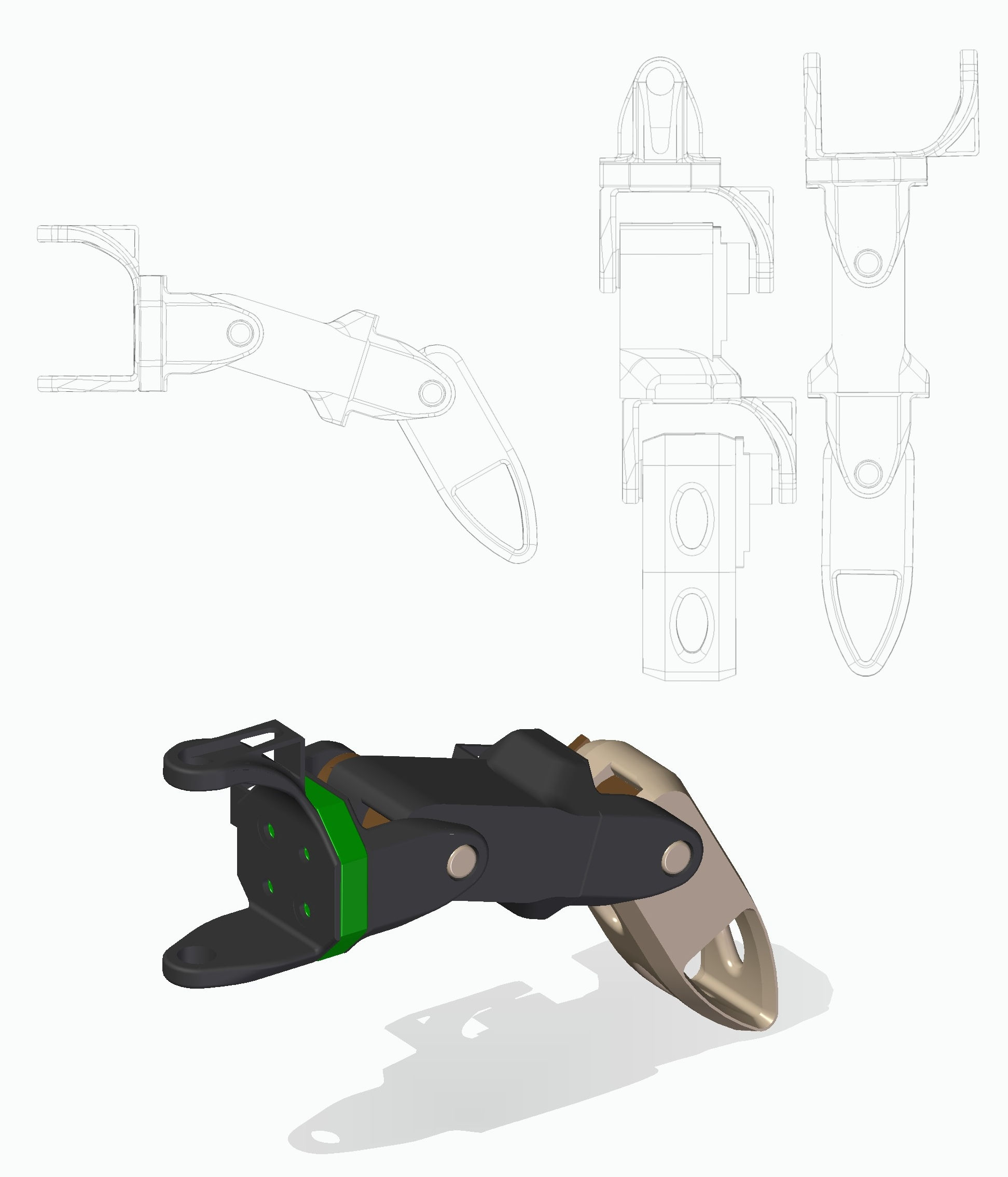}
      \caption{Three joints leg design.}
      \label{fig:3jointleg}
   \end{figure}

As the servos have a $180^\circ$ range, the links should be wide enough to allow for this motion. We designed them so that they can be attached to the next one in different positions. In Fig.~\ref{fig:2jointleg} we show the standard spider leg with 2 joint and in Fig.~\ref{fig:3jointleg} the standard spider leg with 3 joints.

\begin{figure}[!ht]
      \centering
      \includegraphics[width=\linewidth]{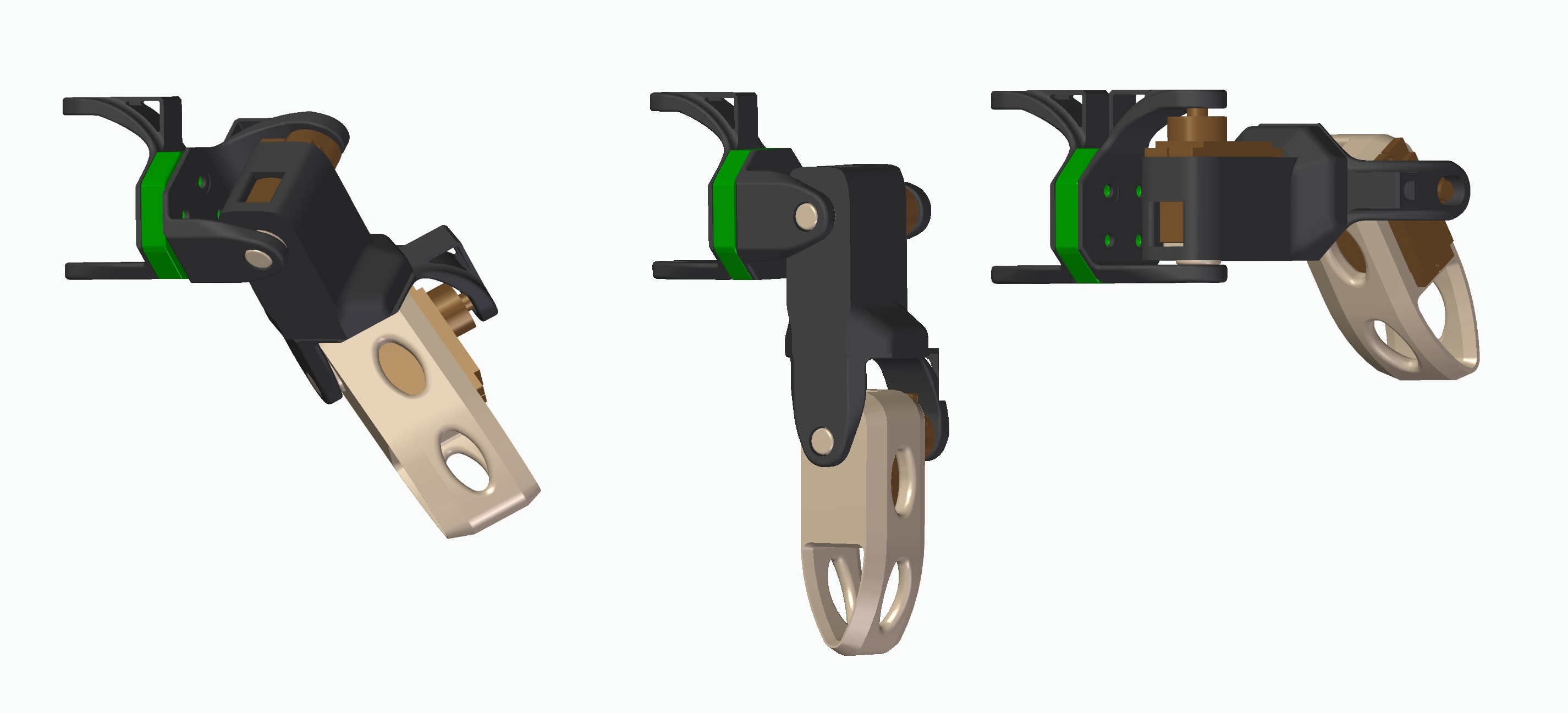}   \caption{Three possible configurations for the leg.}
      \label{leg_configs}
   \end{figure}

   As mentioned in Section \ref{sec:mechanics}, the multiple configurations of the legs (see Fig.~\ref{leg_configs}) give a high value to the LoCoQuad design, since the user can decide or change easily the structure of the robot. In the design process this changed the way the servo horns were placed and how the links need the hexagon to fit more than one configuration.

\section{COST}


Table \ref{tab:locoquadcost} details the component list and unit prices. The prices without brackets correspond to the basic configuration (2 joints per leg) and the prices between brackets to the alternative configuration (3 joints per leg). The cost of the alternative LoCoQuad configuration increases by 15USD due to extra 3D printed parts and the 4 additional servos.

\section{CONCLUSIONS}

LoCoQuad is a novel design for an arachnoid-like quadruped robot. Our main contribution is its low cost, 150USD, which is the lowest for existing legged and wheeled robots of similar technical specifications. This makes it affordable for a wide spectrum of robotic users, which may open new possibilities for research and education.

The platform is open-source, being the mechanical and electrical designs and the code that we developed available online. We believe this is essential for the platform sustainability in the long term and to allow the flexibility that some users might require. We also think that the possibility of building the platform from scratch is very relevant for education.

Our design has also a set of very relevant features. The computational unit is a Raspberry Pi, which combines high computational capabilities and flexibility for programming and interfacing with sensors and actuators. The electrical components are carefully designed for a decent autonomy, and the mechanical parts for a high degree of stability and flexibility for partial re-design.

\section{DISCUSSION AND FUTURE WORK}

\begin{table}[!t]
\caption{Detailed cost for the LoCoQuad platform}
\label{tab:locoquadcost}
\begin{center}
\begin{tabular}{|l c|c|}
\hline
\textbf{Concept} & \textbf{Units} & \textbf{Price[\$]}\\
\hline
\hline
3D parts & $\times 1$ &10 (13)\\
\hline
Raspberry Pi 3 Model B & $\times 1$ &35\\
\hline
Raspberry Pi Camera Module v2.1 & $\times 1$ &30\\
\hline
Micro SD Card 32GB & $\times 1$  &10 \\
\hline
PCA9685 Board 12 PWM channels I2C & $\times 1$  &15\\
\hline
MG90S servos 18Ncm nominal torque & $\times 8 \ (12)$ & 24 (36)\\
\hline
18650 LiPo battery 3000mAh 20C& $\times 2$  &6\\
\hline
2P 18650 LiPo battery holder & $\times 1$  &1 \\
\hline
LiPo charger TP4056 at 1A & $\times 1$  &1\\
\hline
Ultrasonic sensors HC-SR04 & $\times 1$  &1\\
\hline
IMU unit GY-521 3Accel + 3Gyro  & $\times 1$  &4\\
\hline
5V piezoelectric buzzer & $\times 1$  &1\\
\hline
DCDC boost converter 3.2-5V to 5V 15W & $\times 2$  &6\\
\hline
Toggle Switch ON-OFF & $\times 1$ &1\\
\hline
M3 M2.5 M2 screws kit & $\times 1$ &5\\
\hline
\hline
\textbf{TOTAL} &  & \textbf{150 (165)}\\
\hline
\end{tabular}
\end{center}
\end{table}

In our opinion, the LoCoQuad platform that we designed has all the necessary features to be a useful tool for robotics research and education. One of the most critical aspect for this to happen is the creation of a community, and part of our future efforts will be directed to that. We already released our designs, and we plan firstly to advertise and provide support individually to interest groups, and secondly to organize public events as we gather a first group of users. We will continue developing the platform, mainly developing new software for extending the LoCoQuad capabilities, and developing alternative mechanical designs to demonstrate its flexibility.

\addtolength{\textheight}{-5.5cm}   

{\bibliographystyle{ieee}
\bibliography{biblio.bib}
}

\end{document}